\documentclass[12pt]{article}

\usepackage{sbc-template}
\usepackage{graphicx,url}
\usepackage[utf8]{inputenc}
\usepackage[brazil]{babel}
\usepackage{amsthm}
\usepackage{amssymb}
\usepackage{amsmath}

\newcommand{\Xmat}{\mathbf{X}}
\newcommand{\Ahat}{\hat{A}}
\newcommand{\candidates}{\mathcal{C}}
\newcommand{\atomset}{\mathcal{S}}
\newcommand{\predd}{\mathrm{P}_{k,j}}
\newtheorem{definition}{Definition}
\usepackage{amsmath}
\DeclareMathOperator{\margin}{margin}
\usepackage{marvosym}

\title{From Abductive Explanations to Global Logical Rules for Node Classification in SGCs}

\author{Bryan Lima Cavalcante, Thiago Alves Rocha}

\address{
Instituto Federal do Ceará (IFCE)\\
Brazil
\email{bryan.lima.cavalcante07@aluno.ifce.edu.br}
\email{thiago.alves@ifce.edu.br \Letter}
}

\begin{document} 

\maketitle

\begin{abstract}
Graph Neural Networks (GNNs) have achieved remarkable performance in node classification tasks, motivating growing interest in methods capable of explaining their predictions. Recent logic-based approaches, such as LogicXGNN, derive global logical rules for Graph Neural Networks (GNNs) from collections of explanatory subgraphs. While informative, these subgraphs may contain redundant structural information that is specific to individual nodes, potentially limiting the generality of the extracted rules. In this work, we propose a logic-based framework for node classification in Simple Graph Convolution (SGC) networks that uses minimal abductive explanations as an intermediate representation for rule extraction. For each node, we compute a minimal set of node-feature pairs sufficient to preserve the predicted class. These explanations are then used to train decision trees from which global logical rules are extracted. Experiments on benchmark datasets show that the proposed framework produces compact global rules while maintaining high fidelity to the original SGC model.
\end{abstract}

\section{Introduction}\label{sec:introduction}

Graph Neural Networks (GNNs) have become the standard tool for node classification on attributed graphs, with applications ranging from document categorization in citation networks to entity classification in knowledge graphs \cite{ju2024comprehensive,hu2020ogb}. As these models inform decisions in scientific and operational pipelines, explaining individual node predictions and characterising overall model behaviour has become a central concern. Recent surveys organise this literature along two axes: instance-level methods, which explain individual predictions, and model-level methods, which characterise global behaviour~\cite{yuan2022explainability}.

A recent line of work approaches this problem through symbolic logic. Methods such as LogicXGNN~\cite{logicxgnn} derive global logical rules for GNNs by characterizing each node through two complementary predicates. The first is a structural component, obtained from a Weisfeiler--Lehman hash of the node's local multi-hop receptive field. The second is an embedding component, obtained by binarizing a small set of informative dimensions of the learned node representation against thresholds; the dimensions and thresholds are selected by a decision tree fit on the graph-level mean-pooled embeddings against the model's predictions. Correctly classified instances are then encoded as binary vectors over these predicates, and a decision tree trained against the model's prediction distills a per-class set of logical rules, each a disjunction of conjunctions of predicates. A follow-up grounding step links each predicate back to the input by collecting representative subgraphs, namely the local neighborhoods of nodes that activate it; a per-predicate decision tree is then fitted over a canonical, structure-aware concatenation of the node features within those subgraphs, yielding grounding rules that connect predicates to input features. However, the main exposition and experiments of this line target graph classification, with reported baselines such as GLGExplainer and GraphTrail~\cite{glgexplainer,graphtrail} evaluated in the same setting. LogicXGNN sketches an extension to node-level tasks but does not develop or evaluate it. As a consequence, rule-based global explanation for node classification remains comparatively underexplored, and the predicates used by this family carry no formal sufficiency guarantee at the node level.

Simple Graph Convolution (SGC)~\cite{wu2019simplifying} is a natural target for rule-based explanation in this setting. SGC matches the performance of GCN and other state-of-the-art graph neural networks on standard node classification benchmarks despite removing the nonlinearities between message-passing layers and collapsing the resulting transformations into a single linear classifier applied to a fixed feature propagation. The logits of an SGC are therefore a linear function of node features. This property has been used by recent work that distils a nonlinear GNN into an SGC and extracts node-level explanations from it rather than from the original model~\cite{pereira2022convexplainer, pereira2023dnx}; by analogy, explanation pipelines built around SGC can in principle be applied to other node-classification GNNs by first composing with a distillation step, at the cost of an additional approximation.

A parallel line of work uses abductive explanations (AXps)~\cite{ignatiev2019abduction, marquessilva2022delivering}. An AXp is a minimal subset of input feature assignments that, together with the model, entails the predicted class: once this subset is fixed, the model's output does not change regardless of the remaining inputs. AXps therefore provide a formal sufficiency and irredundancy guarantee for the prediction, and, for classifiers that are linear in the inputs being explained, an AXp can be computed in polynomial time rather than via combinatorial search~\cite{marquessilva2020naivebayes}. However, these methods have not been directly applied to GNNs.

To address the limitations of these previous lines of work, we propose AXSGC(Abduction-based Explanations for SGC), a logic-based method for node classification in SGCs. For each node, we extract an AXp at the node-feature level from the SGC classifier and use these AXps as the intermediate representation from which global rules are derived. Rule extraction is applied to the set of node-feature pairs that determine each node's prediction, whose irredundancy is guaranteed by the AXp formulation. We compute these AXps in polynomial time by exploiting the linearity of the SGC score.

Overall, our method operates in three stages. First, for each node, we compute a subset of node-feature pairs that by itself is sufficient to fix the SGC prediction. Second, the resulting sets are encoded as vectors over distance-indexed predicates of the form ''feature of a node at a given hop distance from the target node'', which abstract away the identity of the contributing neighbor while preserving its distance from the target node. These vectors, paired with the SGC prediction as a label, are used to train a decision tree from whose paths the per-class rules are read off. Third, global logical rules are read directly from the root-to-leaf paths of the trees, with each leaf labelled by the class it predicts and each path read as a conjunction of distance-indexed predicates.

Each AXp therefore acts as a bridge between the local and global stages: it is the Node-feature AXp (NF-AXp) returned for an individual node, and at the same time the input from which the global per-class rules are induced.

We evaluate AXSGC on benchmark node classification datasets and report intrinsic fidelity to the SGC model alongside LogicXGNN~\cite{logicxgnn} as a representative recent rule-based global explanation method for GNNs. Our experiments report the
size of AXps, the size of the resulting per-class rules, and the intrinsic fidelity of the rules to the SGC model across the benchmarks considered. Across these benchmarks, our method achieves fidelity to the SGC up to 30.2\% over LogicXGNN and extracting up to 83.8\% fewer rules.

\section{Preliminaries}\label{sec:preliminaries}

This section fixes the notation used in the remainder of the paper. We first recall the SGC classifier and isolate the property that makes it amenable to formal explanation. We then recall the abstract notion of abductive explanation from formal explainability, which serves as the conceptual foundation for the method developed in the next section.

\subsection{Simple Graph Convolution}
\label{sec:prelim-sgc}

Let $G = (V, E)$ be an undirected graph with $n = |V|$ nodes, adjacency matrix $A \in \{0,1\}^{n \times n}$, and node-feature matrix $\Xmat \in \mathbb{R}^{n \times d}$, where $d$ is the number of node features and $X_{u, j}$ is the value of feature $j$ at node $u$. In most datasets considered in this work, node attributes are represented as binary bag-of-words vectors. Hence, $X_{u,j} \in \{0,1\}$, where $X_{u,j}=1$ indicates that term $j$ occurs in the document associated with node $u$, and $X_{u,j}=0$ indicates its absence. Then, throughout the paper, we assume Boolean node features. Let $I \in \mathbb{R}^{n \times n}$ be the identity matrix. The augmented adjacency $A + I$ adds a self-loop to every node, so each node aggregates its own features alongside those of its neighbours. Let $\tilde{D}$ be the diagonal degree matrix of $A + I$, with $\tilde{D}_{ii} = \sum_{k=1}^{n} (A + I)_{ik}$, where $\tilde{D}_{ii}$ is the degree of node $i$ in the augmented graph. The symmetric normalisation by $\tilde{D}^{-1/2}$ on both sides bounds the spectrum of the propagation operator, defined as
\begin{equation}
\Ahat \;=\; \tilde{D}^{-1/2}\,(A + I)\,\tilde{D}^{-1/2}.
\label{eq:ahat}
\end{equation}
The entry $\Ahat_{v, u}$ encodes the one-step weighted influence of node $u$ on node $v$ after self-loop augmentation and symmetric normalisation.

Let $\Theta \in \mathbb{R}^{d \times C}$ be the learned weight matrix, where $C$ is the number of classes and each column $\Theta_{:, c} \in \mathbb{R}^{d}$ is the linear classifier for class $c$, with $\Theta_{j, c}$ the weight assigned to feature $j$ when scoring class $c$. Let $K \in \mathbb{N}$ be the number of times feature propagation is applied; $K$ plays the same role as the number of message-passing layers in a GCN, controlling how far information can travel before classification. The score matrix is given by
\begin{equation}
\hat{Y} \;=\; \mathrm{softmax}\bigl(\Ahat^{K}\,\Xmat\,\Theta\bigr),
\label{eq:sgc-pred}
\end{equation}
with row-wise softmax, where row $v$ of $\hat{Y}$ is the predicted class distribution at node $v$. The predicted class of node $v$ is
\begin{equation}
c^*(v) \;=\; \arg\max_{c \in \{1, \dots, C\}} \bigl(\Ahat^{K}\,\Xmat\,\Theta\bigr)_{v, c},
\label{eq:sgc-argmax}
\end{equation}
where $c^*(v) $ is the class SGC assigns to node $v$.

\subsection{Abductive Explanations}
\label{sec:prelim-axp}

An abductive explanation (AXp)~\cite{ignatiev2019abduction, marquessilva2022delivering} is a minimal set of input assignments whose values determine the model's prediction.

\begin{definition}[Abductive explanation]
\label{def:axp-abstract}
Let $\phi : \mathcal{X} \to \mathcal{Y}$ be a classifier whose feature space factors as $\mathcal{X} = \prod_{i \in F} \mathcal{X}_i$, where $F$ is a finite index set of features and $\mathcal{X}_i$ is the domain of feature $i$, and let $x \in \mathcal{X}$ be an input with prediction $y = \phi(x)$. A subset $H \subseteq F$ is an \emph{abductive explanation} (AXp) for $\phi(x) = y$ if (a) $\phi(x') = y$ for every $x' \in \mathcal{X}$ with $x'_i = x_i$ for all $i \in H$, and (b) no proper subset of $H$ satisfies (a).
\end{definition}

In Definition~\ref{def:axp-abstract}, $\phi$ is the classifier whose prediction is being explained, $F$ is the set of explainable inputs, $\mathcal{X}_i$ is the domain of input $i$, $x$ is the observed input, $y = \phi(x)$ is the prediction at $x$, $H$ is the subset of inputs the explanation holds fixed, and $x'$ ranges over all alternative inputs that agree with $x$ on $H$. Clause (a) says that as long as the values on $H$ are pinned, no perturbation of the remaining inputs $F \setminus H$ can change the prediction. Clause (b) identifies the inputs the model actually relies on rather than a superset that also happens to be sufficient: without (b), $H = F$ would trivially satisfy (a).

\section{AXps for Selecting Node-Feature Pairs}\label{sec:method}

The construction developed in this section turns the sufficiency and minimality conditions of Definition~\ref{def:axp-abstract} into a concrete procedure for explaining SGC node predictions. We proceed in two steps. Subsection~\ref{sec:method-linear} reformulates the SGC classification rule as a system of linear inequalities over node-feature pairs, exposing the structure that the AXp computation exploits. Subsection~\ref{sec:method-local} adepts Definition~\ref{def:axp-abstract} to a node $v$ of an SGC classifier and gives a greedy deletion procedure that returns an AXp at node-feature using a number of model queries linear in the size of the candidate set.

\subsection{Linear inequalities for SGC predictions}
\label{sec:method-linear}

Fix $\Ahat$ and $\Theta$. The entry $(\Ahat^{K})_{v, u}$ is the total weight with which the feature vector of $u$ enters the score at $v$ after $K$ propagation steps; it is nonzero exactly when $u$ is reachable from $v$ in at most $K$ hops on the augmented graph. This entry depends on $u$ only through structural quantities of the augmented graph (degrees and the number of walks of length up to $K$ from $v$ to $u$), and in particular it is zero whenever the shortest-path distance from $v$ to $u$ exceeds $K$. We write $\mathcal{N}_K(v) := \{ u \in V : (\Ahat^{K})_{v, u} \neq 0 \}$ for the $K$-hop receptive field of $v$ and note that $v \in \mathcal{N}_K(v)$. Only feature values at nodes in $\mathcal{N}_K(v)$ can change the score at $v$, so they are the only candidates for an explanation of $c^*(v)$.

Expanding $\Ahat^{K}\,\Xmat$ entrywise as $(\Ahat^{K}\,\Xmat)_{v, j} = \sum_{u \in V} (\Ahat^{K})_{v, u}\,X_{u, j}$ and contracting with $\Theta$, the unnormalised score at $v$ for class $c$ decomposes as a linear combination over node-feature pairs:
\begin{equation}
(\Ahat^{K}\,\Xmat\,\Theta)_{v, c}
\;=\; \sum_{(u, j) \in \mathcal{N}_K(v) \times \{1, \dots, d\}} \alpha^{v,c}_{u,j}\,X_{u, j},
\qquad
\alpha^{v, c}_{u, j} := (\Ahat^{K})_{v, u}\,\Theta_{j, c}.
\label{eq:sgc-linear}
\end{equation}
The coefficient $\alpha^{v, c}_{u, j}$ is the fixed weight with which the Boolean input $X_{u, j}$ enters the unnormalised score of class $c$ at $v$. It factors into a structural term $(\Ahat^{K})_{v, u}$, recording how strongly $u$ reaches $v$, and a semantic term $\Theta_{j, c}$, recording how strongly feature $j$ votes for class $c$. The coefficients depend only on $\Ahat$ and $\Theta$, not on the input features, so they can be precomputed once per node and reused for every explanation query each $v$. Softmax is strictly monotone, so the predicted class is determined by the argmax of the unnormalised scores. The prediction $c^*(v) = c$ therefore holds iff, for every competing class $c' \neq c$, the linear score induced by column $\Theta_{:, c}$ exceeds the linear score induced by column $\Theta_{:, c'}$ evaluated on the same feature values $X_{u, j}$ over the receptive field. The classification rule at $v$ is thus a system of $C - 1$ linear inequalities in the Boolean inputs $\{X_{u, j}\}_{(u, j) \in \mathcal{N}_K(v) \times \{1, \dots, d\}}$:
\begin{equation}
\sum_{(u, j) \in \mathcal{N}_K(v) \times \{1, \dots, d\}} (\alpha^{v, c}_{u, j} - \alpha^{v, c'}_{u, j})\, X_{u, j} \;>\; 0, \qquad \forall\, c' \neq c,
\label{eq:linear-ineq}
\end{equation}
with margin coefficients $\alpha^{v, c}_{u, j} - \alpha^{v, c'}_{u, j}$ fixed by the model. We take $c^*(v)$ throughout the paper as the prediction to be explained.

\subsection{NF-AXps for SGC}
\label{sec:method-local}

Adapting Definition~\ref{def:axp-abstract} for SGC, the components correspond as follows: $\phi$ is the SGC argmax classifier of Eq.~\eqref{eq:sgc-argmax}, $F = \mathcal{N}_K(v) \times \{1, \dots, d\}$ is the set of node-feature pairs inside the receptive field of $v$, $\mathcal{X} = \{0, 1\}^{F}$ is the Boolean cube indexed by these node-feature pairs, and $\mathcal{Y} = \{1, \dots, C\}$ is the set of class labels. The observed input $x$ is the current Boolean configuration on $F$, and $y = c^*(v)$ is the predicted class. A subset $H \subseteq F$ is an AXp at node-feature level for $c^*(v) = c$ iff fixing $X_{u, j} = x_{u, j}$ on $H$ preserves, for every $c' \neq c$, the column-$c$-vs-column-$c'$ inequality derived from~\eqref{eq:sgc-linear}, for all $\{0, 1\}$-assignments of the remaining entries in $F \setminus H$. AXp sufficiency is thus a single linear inequality universally quantified over Boolean completions of $F \setminus H$, with coefficients $\alpha^{v, c}_{u, j}$ fixed by the model.

The natural unit of explanation is therefore a node-feature pair, rather than a whole node or a whole feature. As mentioned in Section~\ref{sec:prelim-sgc}, most datasets considered in our experiments represent node attributes as binary bag-of-words vectors, where $X_{u,j}=1$ indicates the presence of term $j$ at node $u$. We therefore adopt a presence-based explanation semantics. Following robust explanation approaches for text classification, explanations are defined in terms of subsets of the words present in the input~\cite{ijcai2021p0366}. Accordingly, we consider as explanatory candidates only node-feature pairs $(u,j)$ for which the corresponding term is present, i.e., $X_{u,j}=1$. The features that can possibly matter for the prediction at $v$ are therefore exactly those whose pair $(u, j)$ is both reachable and with $X_{u,j} = 1$. We collect these into the candidate set
\begin{equation}
\candidates_v \;=\; \bigl\{(u, j) \,:\, u \in \mathcal{N}_K(v),\; X_{u, j} = 1 \bigr\},
\label{eq:candidates}
\end{equation}
which serves as the search space for the NF-AXp of $v$. We assume: only pairs that are currently switched on $X_{u,j}=1$ can be ``removed'' by setting the corresponding entry to zero, and pairs that are switched off $X_{u,j}=0$ already contribute zero to the score under that baseline.

Let $c^* = c^*(v)$ denote the SGC prediction at $v$. We want a subset $\atomset_v \subseteq \candidates_v$ such that no Boolean completion of the remaining entries of $\Xmat$ on the receptive field, restricted to $\{0, 1\}$ on the indices outside $\atomset_v$ and inside $\candidates_v$, can flip the predicted class. Because the SGC score is linear in $\Xmat$, the worst-case completion is computable explicitly. For every rival class $c' \neq c^*$, define the contribution of a single pair to the margin of $c^*$ over $c'$,
\begin{equation*}
\delta_{u, j, c'} \;=\; (\Ahat^K)_{v, u}\,\bigl(\Theta_{j, c^*} - \Theta_{j, c'}\bigr),
\end{equation*}
and the worst-case margin attainable under a fixed $\atomset \subseteq \candidates_v$ as
\begin{equation}
\margin(c', \atomset) \;=\; \sum_{(u, j) \in \atomset} \delta_{u, j, c'} \;+\; \sum_{(u, j) \in \candidates_v \setminus \atomset} \min\{0,\, \delta_{u, j, c'}\}.
\label{eq:worstcase}
\end{equation}
The two sums admit a direct worst-case reading. Pairs in $\atomset$ are held at their observed value $X_{u, j} = 1$ by the explanation and therefore contribute exactly $\delta_{u, j, c'}$. Pairs outside $\atomset$ can take either Boolean value; the completion that minimises the margin sets $X_{u, j} = 0$ when $\delta_{u, j, c'} > 0$, removing a contribution that would otherwise favour $c^*$, and keeps $X_{u, j} = 1$ when $\delta_{u, j, c'} < 0$, adding a contribution that reduces $c^*$. The minimum operator records exactly this worst-case assignment. Specialising Definition~\ref{def:axp-abstract} to the SGC then yields an explicit sufficiency condition.

We call the AXp for SGC a node-feature AXp (NF-AXp), since it ranges over node-feature pairs $(u, j) \in \mathcal{C}_v$ rather than single feature indices.

\begin{definition}[NF-AXp for SGC at $v$]
\label{def:axp-sgc}
A set $\atomset_v \subseteq \candidates_v$ is a (node-feature) abductive explanation for $c^*(v)$ if $\margin(c', \atomset_v) \geq 0$ for every $c' \neq c^*$. More over, no proper subset of $\atomset_v$ satisfies the same condition.
\end{definition}

A NF-AXp $\atomset_v$ is the analogue, for SGC node classification, of the AXp of Definition~\ref{def:axp-abstract}: a irredundant collection of present node-feature pairs whose value alone, regardless of how every other entry of $\Xmat$ is set within $\{0, 1\}$, already pins the prediction to $c^*$. Equivalently, $\atomset_v$ is a robustness certificate against Boolean erasures on its complement. The full candidate set $\candidates_v$ is trivially sufficient: substituting $\atomset_v = \candidates_v$ into Eq.~\eqref{eq:worstcase} reduces the second sum to zero and recovers the actual margin computed by the model on the observed input, which is non-negative by the definition of $c^*$. So $\candidates_v$ is sufficient but not subset-minimal, and the question is how to compress it.

We obtain a NF-AXp $\atomset_v$ by greedy deletion~\cite{ignatiev2019abduction}. The procedure starts from $\atomset_v \leftarrow \candidates_v$ and processes the elements of $\candidates_v$ in a fixed deterministic order. For each pair $(u, j)$ in turn, it tentatively removes the pair from $\atomset_v$ and recomputes the worst-case margin in Eq.~\eqref{eq:worstcase}; if $\margin(c', \atomset_v \setminus \{(u, j)\}) \geq 0$ holds for every rival class, the pair is dropped permanently, otherwise it is restored. Because the dependence of $\margin$ on $\atomset$ is monotone, in the sense that removing a pair can only weaken the margin, a pair that fails the test once cannot become removable later, so a single linear pass suffices. The result is a NF-AXp at a cost of $|\candidates_v|$ margin checks per node, each of which is itself a sum over $|\candidates_v|$ contributions and so runs in time linear in the candidate set. The total cost of an explanation at $v$ is therefore $\mathcal{O}\bigl((C-1)\,|\candidates_v|^{2}\bigr)$ arithmetic operations.

By construction, the returned $\atomset_v$ inherits the two formal properties of Definition~\ref{def:axp-sgc}. Sufficiency holds because the last accepted state of $\atomset_v$ passes the margin test against every rival; subset-minimality holds because every pair in the final $\atomset_v$ was tested and could not be removed. Both properties are exact, in the strong sense that they hold for the original SGC decision function. These per-node sets are the NF-AXps of AXSGC, and they are the raw material for the global step.

\section{Extracting global rules from AXps}\label{sec:extraction}

We now lift the per-node NF-AXps produced in Section~\ref{sec:method} into a small set of global logical rules. We proceed in two steps. Subsection~\ref{sec:method-predicates} aggregates the per-node sets into a single propositional representation by replacing concrete node identities with distance-indexed predicates over features, producing a Boolean matrix that summarises the entire dataset. Subsection~\ref{sec:method-rules} reads global per-class rules out of this matrix by fitting one shallow decision tree per class against the SGC prediction and converting its root-to-leaf paths into DNF rules whose agreement with the model is measured by fidelity.

\subsection{Distance-indexed predicates and the explanation matrix}
\label{sec:method-predicates}

The NF-AXps $\{\atomset_v\}_{v \in V}$ are detailed certificates but they are written in terms of concrete neighbour identities. Two nodes with structurally similar receptive fields can produce explanations that share no pair $(u, j)$ at all, simply because their neighbours $u$ are different nodes. To generalise across the graph we abstract over node identity while keeping the hop distance at which a contribution arrives. Concretely, for every feature index $j \in \{1, \dots, d\}$ and every hop distance $k \in \{0, 1, \dots, K\}$ we define a Boolean predicate over nodes:
\begin{equation}
\predd(v) \;=\; 1 \;\iff\; \exists\, u \in V : \mathrm{dist}_G(v, u) = k \;\text{and}\; (u, j) \in \atomset_v
\label{eq:predicate}
\end{equation}
where $\mathrm{dist}_G$ is the unweighted shortest-path distance in $G$. In words, $\predd(v)$ asserts that ''feature $j$ appears in the NF-AXp of $v$ through at least one node located exactly $k$ hops away from $v$''. The hop truncation $k \leq K$ is consistent with the SGC receptive field: pairs at greater distance have $(\Ahat^K)_{v, u} = 0$ and therefore cannot appear in any $\atomset_v$.

Two design choices in Eq.~\eqref{eq:predicate} deserve explicit comment. First, the predicate aggregates over neighbours by existential quantification, which is what allows nodes with isomorphic-but-not-identical neighbourhoods to share predicates. Second, the predicate is gated by membership in $\atomset_v$, it fires only when the active occurrence was actually used by the abductive explanation. This second condition is what differentiates the construction from a generic ``bag of features per hop'' encoding, since the predicate selects exactly the contributions that survived the procedure in Subsection~\ref{sec:method-local}.

Collecting the predicates yields a global representation of the dataset. We index columns by pairs $(k, j)$ in lexicographic order and define the explanation matrix
\begin{equation}
M \in \{0, 1\}^{n \times (K + 1)\,d}, \qquad M_{v,\,(k,j)} \;=\; \predd(v).
\label{eq:matrix}
\end{equation}
Each row of $M$ summarises the NF-AXp of a single node in the predicate vocabulary; columns correspond to elementary structural-semantic events of the form ''feature $j$ in a node at hop $k$''. The width $(K+1)\,d$ is bounded a priori by the model and is independent of how the receptive fields of different nodes overlap. The matrix is sparse in practice, since most $\atomset_v$ are small relative to $d$, and it is the only object the global stage of AXSGC needs to consult.

\subsection{Global rule extraction}
\label{sec:method-rules}

The matrix $M$ from Subsection~\ref{sec:method-predicates} reduces global explanation to a tabular learning problem on Boolean inputs, where each input is a node and the target is the SGC prediction at that node. We train a single shallow decision tree $T$ on $M$ with the multi-class target $c^*(v) \in \{1, \dots, C\}$. Crucially, the supervision signal is the SGC prediction and not the dataset label: the tree models the behaviour of the SGC classifier on the predicate vocabulary, so any disagreement between $T$ and $c^*$ is a deficiency of the rules and not of the model, and the corresponding gap is what we will later quantify as fidelity. Splits are scored by Gini impurity and the depth of $T$ is fixed a priori.

The tree $T$ partitions the node set according to short conjunctions of predicate tests. Every internal node of $T$ branches on a single column of $M$, so each root-to-leaf path corresponds to a conjunction of literals of the form $\predd(v) = 0$ or $\predd(v) = 1$, and the leaf at the end of the path carries a single class label $c \in \{1, \dots, C\}$. Reading the literals along a path that terminates at a leaf labelled with class $c$ yields a rule of the form
\begin{equation}
\bigwedge_{(k, j) \in S^{+}} \predd(v) \;\wedge\; \bigwedge_{(k, j) \in S^{-}} \neg\,\predd(v) \;\Longrightarrow\; c^*(v) = c,
\label{eq:rule}
\end{equation}
where $S^{+}$ and $S^{-}$ collect the (hop, feature) tests asserted respectively true and false along the path. Grouping the leaves of $T$ by their class label and taking the disjunction within each group yields one DNF formula per class; the collection of these formulae forms the global rule set returned by AXSGC. The depth of $T$ is bounded by a small constant so that every rule remains short enough to inspect; this also acts as an implicit regulariser against memorising rare predicate combinations. Each literal carries a structural reading: a positive literal $\predd(v) = 1$ asserts that some node at distance $k$ from $v$ contributed feature $j$ to the explanation, while a negative literal asserts the absence of any such contribution at distance exactly $k$ from $v$.

The agreement between the resulting rule set and the SGC model is measured directly. Writing $T(M_{v, \cdot})$ for the class label predicted by $T$ when applied to the row of $M$ indexed by $v$, we define the fidelity of the rule set on a test split $V_{\mathrm{test}} \subseteq V$ as
\begin{equation}
\mathrm{Fidelity} \;=\; \Pr_{v \in V_{\mathrm{test}}}\!\bigl[\,T(M_{v, \cdot}) = c^*(v)\,\bigr],
\label{eq:fidelity}
\end{equation}
i.e.\ the fraction of test nodes on which the rule set reproduces the SGC prediction. Fidelity is the central quantity of the global stage because it isolates the contribution of the rule extractor from the contribution of the underlying classifier: the SGC accuracy with respect to the dataset labels is unaffected by the extraction step, while any disagreement between the rules and $c^*$ is attributable to the abstraction introduced by the predicates and to the capacity limits imposed on the trees.

\section{Experiments}\label{sec:experiments}

The empirical evaluation is organised around three questions raised by the construction of Section~\ref{sec:method}. (Q1) Does the global rule set extracted from the explanation matrix $M$ reproduce the predictions of the underlying SGC, and at what cost in description length? (Q2) How large are the NF-AXps $\atomset_v$ that AXSGC manipulates internally, and how does this scale with dataset properties? (Q3) Are individual rules informative enough to admit a structural reading, and how much of the data does a handful of them cover? Section~\ref{sec:exp-setup} fixes the protocol; Sections~\ref{sec:exp-fidelity}, \ref{sec:exp-sizes} and~\ref{sec:exp-qualitative} address the three questions in turn.

\subsection{Setup}\label{sec:exp-setup}

Four standard node-classification benchmarks are used, spanning the range of regimes AXSGC is intended to cover. BAShapes is a synthetic graph of $700$ nodes built from a Barab\'asi--Albert base augmented with house motifs, with $32$ one-hot degree features and $4$ classes; it serves as a controlled sanity check where ground-truth motifs are known. Cora ($2708$ nodes, $1433$ binary bag-of-words features, $7$ classes) and Citeseer ($3327$ nodes, $3703$ binary bag-of-words features, $6$ classes) are citation networks with natively binary inputs. PubMed ($19717$ nodes, $500$ TF-IDF features, $3$ classes) is an order of magnitude larger and is binarised at zero so that the feature matrix $\Xmat \in \{0,1\}^{n\times d}$ matches the Boolean signature on which the predicates $\predd$.

The base classifier is the SGC of~\cite{wu2019simplifying} with propagation depth $K=2$, trained with learning rate $10^{-2}$, weight decay $5\cdot 10^{-4}$, $200$ epochs and seed $42$. Once $\Ahat$ and the learned weights are fixed, NF-AXps $\atomset_v$ are obtained from the propagated representation $\Ahat^{K}\Xmat$ by the procedure of Section~\ref{sec:method-local}. The predicates $\predd$ of Section~\ref{sec:method-predicates} then populate the explanation matrix $M$, and the global stage of Section~\ref{sec:method-rules} fits a single multi-class decision tree $T$ on $M$ with target $c^*(v)$, using the Gini criterion and maximum depth $12$. Each root-to-leaf path of $T$ yields one rule of the form prescribed by Eq.~\eqref{eq:rule}.

The reference baseline is LogicXGNN~\cite{logicxgnn}, the closest existing rule-based global method for graph neural networks. It is run on the same SGC predictor and the same propagated features, and configured with maximum tree depth $12$ so that the two decision trees have identical capacity. The primary metric is the fidelity of the rule set with respect to the SGC, as defined in Eq.~\eqref{eq:fidelity}, namely the fraction of test nodes on which the class-conditioned DNF rules reproduce $c^*(v)$. Parsimony is measured by the total number of extracted rules. Our implementation is publicly available.\footnote{Anonymous link.}

\subsection{Fidelity and parsimony (Q1)}\label{sec:exp-fidelity}

In addition to fidelity, we report descriptive statistics of the extracted rules: their number, the average length of the antecedents, and their coverage (the fraction of test nodes on which each rule fires). Together, these characterise how compact and how informative the global rules are. Unlike the per-node guarantees of Subsection~\ref{sec:method-local}, the global rules carry no explicit sufficiency certificate. A high fidelity is empirical evidence that the abductive vocabulary captures the SGC decision boundary.

A rule-based explainer is only useful insofar as it faithfully tracks the model it claims to describe, and that faithfulness should be obtained with a rule set small enough to inspect. Table~\ref{tab:fidelity} reports, for each dataset, the SGC test accuracy together with the fidelity and rule count of the two global explainers. The SGC accuracy is included as a reference point so that fidelity is read against the predictive quality of the model being explained, not against ground-truth labels.

\begin{table}[t]
\centering
\caption{Test accuracy of the SGC classifier and fidelity / rule count of the two global rule-based explainers under identical tree depth $12$.}
\footnotesize
\label{tab:fidelity}
\begin{tabular}{|l|c|c|c|c|c|}
\hline
 & & \multicolumn{2}{c|}{AXSGC} & \multicolumn{2}{c|}{LogicXGNN} \\
\cline{3-4}\cline{5-6}
Dataset & SGC acc.\ (\%) & Fidelity (\%) & rules & Fidelity (\%) & rules \\
\hline
BAShapes & 100.00 & 100.00 & 6   & 100.00 & 4   \\
\hline
Cora     &  74.40 &  90.70 & 77  &  78.60 & 318 \\
\hline
Citeseer &  61.70 &  83.50 & 72  &  76.60 & 445 \\
\hline
PubMed   &  99.90 &  99.90 & 31  &  69.70 & 153 \\
\hline
\end{tabular}
\end{table}

On BAShapes both methods reach perfect fidelity, with rule counts of the same order as the number of latent motif classes: both methods simply rediscover those motifs, and AXSGC does so with six rules against four. On Cora, AXSGC reaches $90.7\%$ fidelity with $77$ rules, against $78.6\%$ with $318$ rules for LogicXGNN. Citeseer shows the same qualitative pattern at lower absolute fidelity for both methods, with AXSGC reaching $83.5\%$ fidelity against $76.6\%$ for LogicXGNN, a gain of $6.9$ percentage points while using $83.8\%$ fewer rules ($72$ vs $445$). On PubMed the gap is the largest of the four benchmarks: $99.9\%$ fidelity with $31$ rules against $69.7\%$ with $153$. A plausible reading is that the NF-AXp step produces candidate node-feature pairs tailored to each node before aggregation. A single rule can then subsume many NF-AXps once they share a common minimal core. Decision-tree path enumeration on the raw propagated features, in contrast, does not exploit this redundancy.

\subsection{NF-AXp sizes (Q2)}\label{sec:exp-sizes}

The global rule set is the externally visible artefact, but it is built by aggregating the NF-AXps $\atomset_v$ produced by greedy deletion. Their size controls both the cost of AXSGC and the readability of the intermediate evidence that an inspector might want to query for individual nodes. Table~\ref{tab:sizes} reports the mean, median and maximum of $|\atomset_v|$ over the whole node set of each dataset.

\begin{table}[t]
\centering
\caption{Distribution of the NF-AXp size $|\atomset_v|$ across all nodes of each dataset.}
\footnotesize
\label{tab:sizes}
\begin{tabular}{|l|c|c|c|c|}
\hline
Dataset & $n$ & mean $|\atomset_v|$ & median $|\atomset_v|$ & max $|\atomset_v|$ \\
\hline
BAShapes &   700 &   4.0 &   2 &    31 \\
\hline
Cora     &  2708 &  67.4 &  32 &  2426 \\
\hline
Citeseer &  3327 &  61.7 &  35 &  1238 \\
\hline
PubMed   & 19717 & 629.2 & 292 & 45288 \\
\hline
\end{tabular}
\end{table}

The mean NF-AXp size scales jointly with the receptive-field density induced by $\Ahat^{K}$ and with the feature dimension. BAShapes, whose $32$-dimensional one-hot vector. Because this encoding is extremely sparse after propagation, yields NF-AXps of median size $2$. Cora and Citeseer, which share the bag-of-words regime, lie in the same range despite Citeseer's larger feature dimension. PubMed, whose denser TF-IDF signature activates many coordinates per node after two hops of $\Ahat^{K}\Xmat$, produces the largest NF-AXps.

In all three non-synthetic datasets the median of $|\atomset_v|$ is markedly smaller than the mean, and both are orders of magnitude smaller than the maximum. This long-tailed shape indicates that a small number of high-degree hubs concentrate the cost of explanation, while the typical node admits a substantially shorter sufficient condition than the mean alone would suggest. This is precisely the regime in which the aggregation of the $\atomset_v$ through the matrix $M$ is most beneficial: many short, overlapping NF-AXps collapse into a few global rules, while the rare long ones are absorbed by the same aggregation step.

\begin{table}[t]
\centering
\caption{Mean rule length (number of conditions per rule) of the two global rule-based explainers under identical tree depth $12$.}
\footnotesize
\label{tab:rule-length}
\begin{tabular}{|l|c|c|}
\hline
Dataset & AXSGC & LogicXGNN \\
\hline
BAShapes &  2.50 & 1.75 \\
\hline
Cora     &  8.05 & 9.41 \\
\hline
Citeseer &  9.49 & 9.64 \\
\hline
PubMed   &  4.21 & 7.62 \\
\hline
\end{tabular}
\end{table}

Beyond the size of the intermediate NF-AXps, we also compare the size of the global rules between AXSGC and LogicXGNN. Table~\ref{tab:rule-length} reports the mean number of conditions per rule for both methods on each dataset. On BAShapes, both methods produce very short rules, with LogicXGNN slightly shorter ($1.75$ vs $2.50$). On the citation networks (Cora and Citeseer), the two methods produce rules of comparable length, in the range of $8$--$10$ conditions. On PubMed, AXSGC produces noticeably shorter rules ($4.21$ vs $7.62$), which is consistent with the smaller number of rules it requires (Table~\ref{tab:fidelity}) and with the long-tailed shape of the NF-AXp sizes: a large fraction of nodes admits a short sufficient condition that the global tree can absorb into a compact disjunction.

\subsection{A qualitative example and coverage (Q3)}\label{sec:exp-qualitative}

Fidelity numbers and average sizes do not by themselves indicate whether individual rules carry semantic content. As a concrete illustration, consider the highest-coverage rule extracted on BAShapes, denoted $R_0$ and assigned to class $0$ (the Barab\'asi--Albert base nodes). It fires on $280$ of the $700$ nodes ($40\%$ of the dataset).
\begin{equation*}
R_0:\quad P_{0,31}(v) = 1 \ \wedge\ P_{1,2}(v) = 0 \ \Rightarrow\ c^*(v) = 0.
\end{equation*}
On BAShapes feature $31$ is the highest-degree bin of the one-hot degree encoding and feature $2$ is the degree-two bin, so the rule asserts that a node belongs to the BA base whenever its own zero-hop representation activates the high-degree bin and no node at one hop activates the degree-two bin. The condition is structural rather than feature-driven and matches the way the synthetic graph is generated, which is the kind of correspondence one would hope to recover from a transparent model.

\begin{table}[t]
\centering
\caption{Top-1 and top-10 coverage of AXSGC and LogicXGNN rules per dataset.}
\footnotesize
\label{tab:coverage}
\begin{tabular}{|l|c|c|c|c|c|}
\hline
 & & \multicolumn{2}{c|}{AXSGC} & \multicolumn{2}{c|}{LogicXGNN} \\
\cline{3-4}\cline{5-6}
Dataset & $n$ & top-$1$ & top-$10$ & top-$1$ & top-$10$ \\
\hline
BAShapes &   700 &  280 &   700 ($100.0\%$) &  300 &   700 ($100.0\%$) \\
\hline
Cora     &  2708 &  374 &  2162 ($79.8\%$)  &  232 &   936 ($34.6\%$)  \\
\hline
Citeseer &  3327 &  740 &  2846 ($85.5\%$)  &  125 &   679 ($20.4\%$)  \\
\hline
PubMed   & 19717 & 6783 & 19477 ($98.8\%$)  & 2045 & 10528 ($53.4\%$)  \\
\hline
\end{tabular}
\end{table}

Coverage is also informative on the citation networks, where exhaustive inspection of the rule list is not realistic. Table~\ref{tab:coverage} reports the top-$1$ and top-$10$ coverage for both methods. On BAShapes both reach $100\%$ with their ten top rules. On the larger benchmarks, AXSGC rule sets are markedly more concentrated: the top $10$ rules cover $79.8\%$ of Cora, $85.5\%$ of Citeseer and $98.8\%$ of PubMed, against $34.6\%$, $20.4\%$ and $53.4\%$ for LogicXGNN --- gaps of $2.3\times$, $4.2\times$ and $1.85\times$ respectively. AXSGC therefore concentrates coverage in a small subset of high-coverage rules, whereas LogicXGNN spreads it more thinly across its larger pool, consistent with its higher rule counts and lower fidelity (Table~\ref{tab:fidelity}).

\section{Conclusion}\label{sec:conclusion}

We presented AXSGC, a logic-based framework for explaining SGC node predictions. For each node, AXSGC derives an AXp with an explicit sufficiency margin from the closed form of Simple Graph Convolution. These per-node AXps are then aggregated into a global per-class ruleset via a decision tree over distance-indexed predicates. Across four benchmarks, our method improves fidelity by up to 30.2\% over LogicXGNN while extracting up to 83.8\% fewer rules. AXSGC delivers per-node certified sufficiency and empirically faithful, model-aligned global rules. We plan to extend AXSGC in two directions. First, we will apply it to other GNN architectures by composing it with a distillation step to SGC~\cite{pereira2023dnx}. Second, we will extend our method and the NF-AXps directly to other GNN architectures.

\bibliographystyle{sbc}
\bibliography{sbc-template}

\end{document}